\documentclass[runningheads]{llncs}
\usepackage[utf8]{inputenc}
\usepackage{xcolor}
\usepackage{graphicx}
\usepackage{textcomp}
\usepackage{array}
\newcolumntype{P}[1]{>{\centering\arraybackslash}p{#1}}
\usepackage{lipsum}
\usepackage{url} \usepackage{hyperref}
\usepackage{tabularx} 

\newcommand\qquote[1]{``\emph{#1}"} % permet de faire des citations de texte rapide avec les guillements francaises  

\title{A Novel Skill Modeling Approach: Integrating Vergnaud's Scheme with Cognitive Architectures}%Améliorer la modélisation des compétences : Intégration du schème de Vergnaud dans les architectures cognitives
\titlerunning{Improved skill modelling with cognitive architectures}%{Enhancing modelisation of skill with cognitive architecture}
\author{Antoine LENAT\inst{1,2}, Olivier CHEMINAT\inst{2}, Damien CHABLAT\inst{1}, Camilo CHARRON\inst{1,3}}
\authorrunning{A. LENAT et al.}
\institute{
$^1$Nantes Université, École Centrale Nantes, CNRS, LS2N, UMR 6004 \\
44000 Nantes, France\\
$^2$ Université Rennes 2, Rennes, France\\
$^3$ CETIM, 74 Route de la Jonelière, 44300 Nantes \\
\email{\{Antoine.Lenat, Camilo.Charron, Damien.Chablat\}@ls2n.fr}}
\institute{
$^1$Nantes Université, École Centrale Nantes, CNRS, LS2N, UMR 6004 \\
44000 Nantes, France\\
$^2$ CETIM, 74 Route de la Jonelière, 44300 Nantes \\
$^3$ Université Rennes 2, Rennes, France\\
\email{\{Antoine.Lenat, Damien.Chablat, Camilo.Charron\}@ls2n.fr}, Olivier.Cheminat@cetim.fr}

\date{\today}

\begin{document}
\begin{sloppypar}

\maketitle

\begin{abstract}
Human-machine interaction is increasingly important in industry, and this trend will only intensify with the rise of Industry 5.0. Human operators have skills that need to be adapted when using machines to achieve the best results. It is crucial to highlight the operator's skills and understand how they use and adapt them \cite{lenat_improving_2024}. A rigorous description of these skills is necessary to compare performance with and without robot assistance. Predicate logic, used by Vergnaud within Piaget's scheme concept, offers a promising approach. However, this theory doesn't account for cognitive system constraints, such as the timing of actions, the limitation of cognitive resources, the parallelization of tasks, or the activation of automatic gestures contrary to optimal knowledge. Integrating these constraints is essential for representing agent skills understanding skill transfer between biological and mechanical structures. Cognitive architectures models \cite{albus_rcs_2005} address these needs by describing cognitive structure and can be combined with the scheme for mutual benefit. Welding provides a relevant case study, as it highlights the challenges faced by operators, even highly skilled ones. Welding's complexity stems from the need for constant skill adaptation to variable parameters like part position and process.  This adaptation is crucial, as weld quality, a key factor, is only assessed afterward via destructive testing. Thus, the welder is confronted with a complex perception-decision-action cycle, where the evaluation of the impact of his actions is delayed and where errors are definitive. This dynamic underscores the importance of understanding and modeling the skills of operators.

\keywords{Skills \and Cognitive architectures \and Scheme \and Cognition \and Robotics \and Welding}
\end{abstract}

%%%%%%%%%%%%%%%%%%%%%%%%%%%%%%%%%%%%%%%%%%%%%%%%%%%
\section{Skill Representation : Theoretical Foundation}
%%%%%%%%%%%%%%%%%%%%%%%%%%%%%%%%%%%%%%%%%%%%%%%%%%%
The increasing prevalence of human-robot interaction in modern industry implies to the welder the necessity to adapt there skills without robot to a situation with new technologies. However, this adaptation is often challenging, highlighting the need for robots to adapt to human operators rather than the other way around, thus preventing workflow bottlenecks. Skill, in this context, is a multifaceted concept explored across diverse disciplines, including professional didactic, ergonomics, academic training, cognitive psychology, work psychology, aging psychology, and neuroscience \cite{lenat_improving_2024}. 

Welding is a good example to show the complexity of skill since the quality of the bead cannot be seen directly and must go through destructive testing; specificity called special process, according to aeronautical terminology \cite{en_9100_quality_2018}. Thus, the gratification of success or the penalty for non-compliance occurs with a time delay and not systematically. This bias leads to complex learning in the case of special processes. Default being permanent is making crucial the operator skill during the process.

Describing skills requires models to represent it according to its field of application. Our definition of skill rely on the one given by Coulet \cite{coulet_notion_2011}, namely \qquote{a dynamic organization of activity, mobilized and regulated by a subject to address a given task in a determined situation}. This definition show the proactive regulation imperative to adaptation.

Thus, we can conclude that skill expresses a potential that an individual can realize to perform a task in a situation while accounting for possible variations. Although it is not directly observable, it can be inferred from observable elements, which may take the form of material remnants or even a pause suggesting a moment of reflection, for instance. What an individual is capable of performing already exists in the form of resources and adapts to the situation through retroactive regulation based on the individual's experiences and interpretation of the situation. Thus, skill is organized continuously, dynamically, and is mobilized within a class of situations.

This definition is close to what can be found in Vergnaud's work since Coulet was inspired by him \cite{coulet_organization_2019}. Here, we can highlight the similarity between this definition and Vergnaud's scheme, defined as the \qquote{invariant organization of behavior for a given class of situations} \cite{vergnaud_theory_2009} in his conceptual field theory. Thus, we will consider skill equivalent to the scheme. Based on predicate logic, the scheme is structured around four pillars: inferences (information gathering), action rules (decomposition into elementary tasks), expectations (goals and anticipation to be achieved), and operational invariants (beliefs deemed relevant or irrelevant in a situation). Inferences are the computational aspect of the scheme, linking specific parameters to rules of action (\qquote{if such parameters then such rules of action}) A dynamic situation is defined as a situation that evolves even without human intervention; Hoc \& Amalberti \cite{hoc_cognitive_2007} define it as a situation where \qquote{the human operator only partially controls the technical process or the environment.}  

\begin{figure}[!ht]
    \vspace{-0,3 cm}
    \centering
    \includegraphics[width=0.65\textwidth]{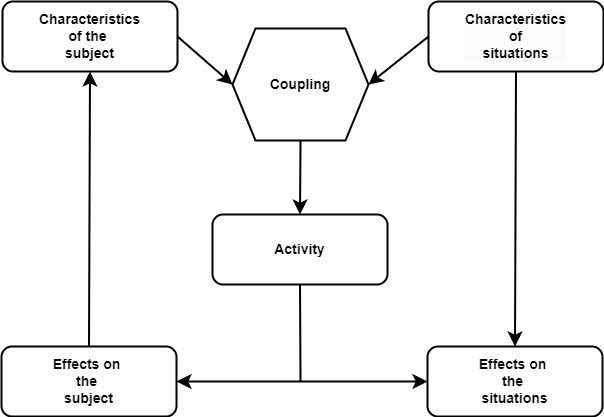}
    \caption{Dual regulation model, extract from \cite{vidal-gomel_competences_2016} (itself from \cite{leplat_regards_1997})}%{Modèle de double régulation, extrait de \cite{vidal-gomel_competences_2016} d'après \cite{leplat_regards_1997}}
    \label{fig:Double Regulation}
    \vspace{-0,3 cm}
\end{figure}

The situation is at the core of skill execution as it determines the operator's degrees of freedom through constraints \cite{vidal-gomel_competences_2016}. It allows the operator to produce an activity (productive task) and to develop themselves by acquiring knowledge (constructive task). However, by interacting with the situation to solve the problem, the operator alters the nature of the situation, creating a form of regulatory loop. In doing so, they also adapt and modify their goal by taking the changes into account, regulating their own activity. This dual regulation process has been described by Leplat \cite{leplat_regards_1997}, as illustrated in figure \ref{fig:Double Regulation}  \cite{rogalski_chap_2019,vidal-gomel_competences_2016}.

This regulation of the subject can be observed in the short term through joy (I have made a good weld bead in a difficult situation) or frustration (I could have done a better one), depending on success or failure, or through the adjustment of actions during the activity. 

Skills are often described using semantic models, but we aim to represent them schematically, taking our inspiration from the diagrams used in control laws familiar to roboticists. Conceptual field theory provides the mathematical precision, particularly with predicate logic, needed to formalize skills in equation form, making a transition from semantic to stochastic skill models. However, schemes do not take into account the functional limitations of human cognition, such as constraints on task parallelization, cycle time delays, and working memory capacity. Other models and theories can be used to fill this gap, such as cognitive architectures, which deal with the general functioning of cognitive processes.

%%%%%%%%%%%%%%%%%%%%%%%%%%%%%%%%%%%%%%%%%%%%%%%%%%%
\section{Cognitive Architectures: A Comparative Overview}
%%%%%%%%%%%%%%%%%%%%%%%%%%%%%%%%%%%%%%%%%%%%%%%%%%%

Developed in the early 1980s, cognitive architectures aim to model cognitive phenomena. The term refers both to representations of cognition and its structure, and how those representations are implemented in computers as software. Focused on the internal properties of the cognitive system, models were created to simulate the functioning of a cognitive system in the form of computational processes designed for artificial intelligence. Two models among the most well-known architectures doing are the SOAR model \cite{laird_soar_2012} (State Operator And Results) and ACT-R model \cite{anderson_integrated_2004} (Adaptive Control of Thought - Rational).

Cognitive architectures can be described as a general and invariant framework for a \textit{superordinate class of situations, essentially a description of the overall functioning of the cognitive system}, as shown by Hoc and Amalberti \cite{hoc1999analyse}. We appreciate the use of class of situation in this description to stay close to Vergnaud theory. However, the definition given by Albus \cite{albus_rcs_2005} can highlight some aspects of cognitive architectures models, therefore a cognitive architecture can be defined as \qquote{the organizational structure of functional processes and the knowledge representations that enable the modeling of cognitive phenomena}. The elements in this definition highlight four key aspects of cognitive architectures: 
\begin{enumerate}
    \item an organizational structure providing a hierarchy, roles, and relationships between modules,
    \item functional processes, or cognitive mechanisms, encompassing all internal operations (perception, interaction with memory, information processing),
    \item a representation of information storage and its forms (scripts, symbols, etc.) to interpret sensory data,
    \item a modeling of cognitive phenomena, including the results of interactions between structure and representations, such as decision-making, comprehension, or language processing.
\end{enumerate}

Another definition is provided by the Institute for Creative Technologies\footnote{\url{https://ict.usc.edu/research/labs-groups/cognitive-architecture/}}, which defines architecture as a \textit{hypothesis about the fixed structures that provide a mind, whether in natural or artificial systems, and how they work together – in conjunction with knowledge and skills embodied within the architecture – to yield intelligent behavior in a diversity of complex environments}. This definition incorporates the same elements as Albus's, but it also highlights the importance of the relationships between the structural elements and the embodied nature of the knowledge integrated within it.

In these definitions, the first step of cognitive phenomena is the perception of the environment, therefore it is needed to describe this interaction between an agent and it's environment. Inspired by the perception-action loop from neuroscience, Laird \cite{laird_standard_2017} proposes a description of a cognitive cycle involving information acquisition from the environment through sensory organs, processing within working memory (including a comparison between the contents of working memory and procedural memory to select an action), and the execution of an action (cognitive or motor). 

Although the scheme is specific to a situation and cognitive architectures pertains to overall functioning, their definitions can bring several comparisons:  
\begin{itemize}
    \item Both the scheme and cognitive architectures emphasize the existence of an organized structure underlying cognitive activity. The scheme represents a structure of actions or thoughts, whereas cognitive architectures represents a structure of processes and representations.
    \item The action rules of the scheme represent the succession of elementary actions, which parallel the functional processes and cognitive phenomena of architectures.
    \item The operational invariants are a way of representing knowledge, elements that are also present in Albus’s definition.
    \item Operational invariants explain the action rules based on inferences. Laird's definition of cognitive cycle describes the selection of an action as a comparison between the content in working memory and the content in long-term memory.
    \item The limitations of the scheme in describing the boundaries of cognitive functioning are addressed by cognitive architectures.
\end{itemize}

Finally, we propose modeling the scheme by drawing inspiration from perception-action loops to represent the regulation loops of skill in block diagrams, incorporating concepts from cognitive architectures. To achieve this, we will explore several major models and see how welding can be encompass within.

%%%%%%%%%%%%%%%%%%%%%%%%%%%%%%%%%%%%%%%%%%%%%%%%%%%
\subsection{Toward a standard model}
%%%%%%%%%%%%%%%%%%%%%%%%%%%%%%%%%%%%%%%%%%%%%%%%%%%

%%%%%%%%%%%%%%%%%%%%%%%%%%%%%%%%%%%%%%
\subsubsection{Laird's SOAR model}
%%%%%%%%%%%%%%%%%%%%%%%%%%%%%%%%%%%%%%

The SOAR model, which stands for State, Operator And Result, was developed by J. Laird and A. Newell in the 1980s and continued to evolve until 2012 \cite{laird_soar_2012}. This architecture provides a framework for describing a cognitive cycle, encompassing a set of cognitive functions such as memory management mechanisms, content representation, and learning models for knowledge acquisition. The cycle incorporate all this functional processes to modeling cognitive phenomena by describing the way an agent interact with environment through perceptions and motor gestures (or actions) while treating information in the memories modules. In the example of welding, we see the weld pool (perception), we use our knowledge to regulate our action and modify our gesture (action) : I'm seeing a welding pool too small, I have to decrease my traveling speed to increase the amount of energy.

\begin{figure}
    \vspace{-0,3 cm}
    \centering
    \includegraphics[width=0.9\linewidth]{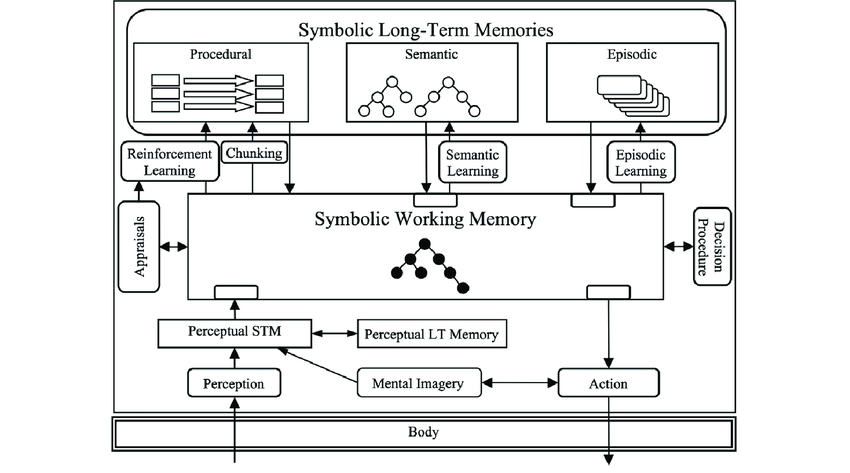}
    \caption{Representative diagram of the SOAR model}%{Schéma représentatif du modèle SOAR}
    \label{fig:SOAR}
    \vspace{-0,3 cm}
\end{figure}

To explain the model, we will describe the cognitive cycle depicted in figure \ref{fig:SOAR}, focusing on its main step, being the interaction between body and environment, treatment of information in working memory thanks to it's access with long term memory. While the model comprises several sub-modules, these can be selectively omitted depending on the specific study, like Laird's standard model shows \cite{laird_standard_2017}. We can then found learning modules (semantic, episodic or reinforcement), decision procedure, chunking or the buffer between perception and working memory. Chunking is a cognitive phenomenon that involves grouping data together to reduce the load on working memory and increase the ability to perceive familiar patterns \cite{chase_minds_1973} \cite{gobet_psychologie_2011}. If we take the case of welding a chunking example could be assembling the color of weld pool, its size, the sound of torch as a whole. The sum of the whole give more information as the sum of the parts.

A cognitive cycle begins with the perception of information from the environment through the body and going to the working memory. 

The memory process allows for the storage and retrieval of content to support the specific functions necessary for working memory, such as learning or managing symbols and relationships. Working memory is a dynamic space using buffers to interact with other modules, accelerating their interactions, similar to random-access memory in computers. Buffers therefore contain the knowledge required for problem-solving, the perception of sensory information or the intermediate results of problem-solving, for example.

Procedural memory contains all the information (knowledge) related to internal or external actions. This may include the choice of action (decision-making) or how to perform it. Laird proposes that procedural memory is based on patterns, similar to ACT-R. Other memories can store facts and concepts; Laird describes them as a graph of symbolic relations.

Learning is seen as the automatic creation of new symbolic structures and a modification of metadata in both long-term memories (declarative and procedural). It also involves the adaptation of perception contents and the motor system (p.13). Learning is therefore based on two paths: the creation of new rules or relation, or reinforcement/adaptation.

If we have to weld a new nuance of material, we use our experience (long term memory) and adapt our gesture according to our knowledge (symbolic representation), accommodation of scheme according to Piaget.

%%%%%%%%%%%%%%%%%%%%%%%%%%%%%%%%%%%%%%
\subsubsection{Anderson's ACT-R model}%{Modèles ACT-R}
%%%%%%%%%%%%%%%%%%%%%%%%%%%%%%%%%%%%%%

Between the 1970s and 1990s, J.R. Anderson developed a series of models of human cognition. The first architecture model was named ACT for \qquote{\textit{Adaptive Control of Thought}}. Later, Anderson sought to add mathematical consistency to the model, which he called a Rational Analysis. The model then became ACT-R during the 1990s. ACT-R continued to evolve until the mid-2010s with the advent of brain imaging.

The fields of application for the ACT-R model are numerous, with more than 700 related articles. The website dedicated to the ACT-R community \cite{act-r_research_group_act-r_2024} provides a relatively comprehensive list of articles by field of application, among which we can cite, language processing, perception and attention, problem-solving and decision-making, learning and memory.

We wish to focus on the functioning of the ACT-R model itself, and will not develop the evolution of the model further here. We know that this model has already been used for the creation of artificial intelligence and software. However, we are not yet at a stage of writing computer code, and we will not describe software management within the context of the ACT-R model here.

ACT-R is a two-level cognitive architecture: symbolic and subsymbolic. The former represents knowledge and reasoning, while the latter handles the dynamic management of knowledge activation and cognitive time management.

\begin{figure}
    \vspace{-0,3 cm}
    \centering
    \includegraphics[width=0.7\linewidth]{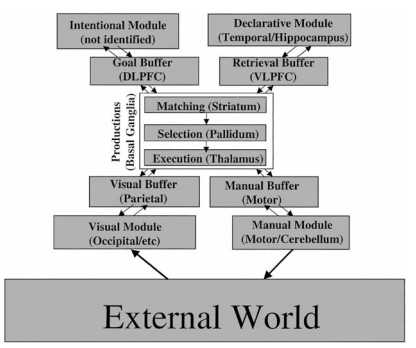}
    \caption{Model ACT-R 5.0 as described by Anderson \cite{anderson_integrated_2004}}%{Modèle ACT-R 5.0 tel que décrit pas Anderson \cite{anderson_integrated_2004}}
    \label{fig:Modèle ACT-R}
    \vspace{-0,3 cm}
\end{figure}

Anderson describes the ACT-R 5.0 model \cite{anderson_integrated_2004} using a diagram presented in Fig \ref{fig:Modèle ACT-R}. As Anderson states in this paper, this is a basic architecture of the model. The theory does not specify exactly how many modules exist, making this model adaptable based on the goals and fields of study.

\begin{figure}[!ht]
    \vspace{-0,3 cm}
        \centering
        \includegraphics[width=0.65\linewidth]{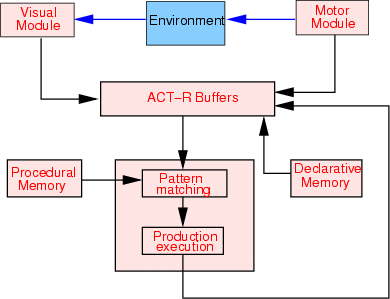}
        \caption[]{Representative diagram of the ACT-R 5.0 model \cite{act-r_research_group_act-r_2024}}.  
        \label{fig:Modèle ACT-R revisité}
    \vspace{-0,3 cm}
\end{figure}

The official website of the ACT-R research group \cite{act-r_research_group_act-r_2024} provides a simplified version of the model, presented in Fig. \ref{fig:Modèle ACT-R revisité}. Where we can notably see buffers, modules and links.

The pattern matching represents association of model, composed of matching, selection and execution. These three steps are encompassed in basal ganglia in the figure \ref{fig:Modèle ACT-R}. The matching rely on the comparison of the ACT-R Buffer, supplied by visual and motor buffer, and the procedural memory. The selection of the optimal action to perform is based on the goal buffer and the state of the situation. Once the action is performed, a loop goes back to the ACT-R Buffer, reminding the dual regulation of activity. However, we could add a link from the buffer to memories modules to depict this notion with more precision.

Anderson describes the production of a single rule to match the current model \qquote{\textit{a single production rule is selected to respond to the current pattern}} [p.1]. This pattern matching recalls the activation and inhibition of schemes described in Pascual Leone's model \cite{arsalidou_constructivist_2016,de_ribaupierre_modeneo-piagetiens_2007}.

%%%%%%%%%%%%%%%%%%%%%%%%%%%%%%%%%%%%%
\subsubsection{Laird's standard model}
%%%%%%%%%%%%%%%%%%%%%%%%%%%%%%%%%%%%%

The standard model proposed by John Laird \cite{laird_standard_2017} is based on a compilation and simplification of multiple models, mainly the ACT-R \cite{anderson_integrated_2004} and SOAR \cite{laird_soar_2012} models (also SIGMA \cite{steunebrink_rethinking_2016}, which is really close to SOAR model and describe the same perception action-loop).

\begin{figure}
    \vspace{-0,3 cm}
    \centering
    \includegraphics[width=0.80\linewidth]{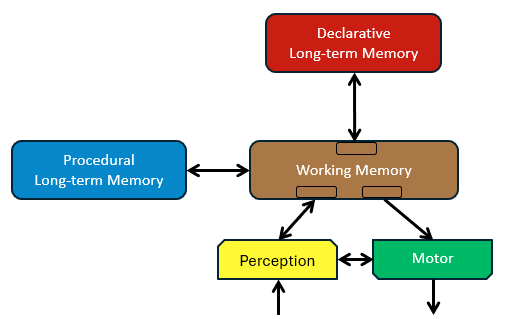}
    \caption{Representative diagram of the standard model \cite{laird_standard_2017}}%{Schéma représentatif du modèle standard \cite{laird_standard_2017}}
    \label{fig:Standard}
    \vspace{-0,7 cm}
\end{figure}

As shown in figure \ref{fig:Standard}, the standard model proposes a synthetic structure with a limited set of modules and their interconnections : perception, movement (motor), working memory, declarative long-term memory, and procedural long-term memory. The cognitive cycle described here is close to SOAR, beginning with body perception, going through working memory and long term memory. However, lots of modules are missing, like the learning (semantic, episodic or reinforcement), chunking or decision procedure. Interaction between perception and motor are also simplify. 

The simplified vision of the standard model is intended to highlight the fundamental mechanisms of the cognitive cycle, with the details of each module being described more precisely by adding sub-modules. Long term memory, such as declarative, is especially used for learning and information retrieval. They rely on the use of buffers in both working memory and declarative memory. 

The standard model does not consider the mind as a set of undifferentiated information \cite{laird_standard_2017}, but rather as a set of independent modules, each with its own functionality. As stated in the article (p.19), this version of the standard model is simplistic and offers little evolution compared to older models, except for the distinction made between procedural and declarative memory within long-term memory, which are seen as independent here.

According to Laird, the goal of a cognitive cycle is to test the contents of working memory and select an action to modify that content (p.17). He goes further and describes it as resulting from procedural memory, which induces the choice to perform an act per cycle (p.20). The modification of the content can be abstract reasoning, the internal simulation of an external action, its execution, or the retrieval of information from long-term memory, for example. Cognitive cycles can occur in parallel between components.

Like the SOAR model, the standard model allows for the description of a cognitive cycle starting with the organism's sensory receptors, providing a set of data to working memory, which will use it as a source of information via its access to long-term memory modules. The sensory data are also connected to the motor module for reflex situations (e.g., if I burn myself, I pull my hand away immediately). Once processed in working memory, an action is carried out, whether cognitive or physical.

%%%%%%%%%%%%%%%%%%%%%%%%%%%%%%%%%%%%%%
\subsection{Complementary Perspectives: Further Models of Cognition}
%%%%%%%%%%%%%%%%%%%%%%%%%%%%%%%%%%%%%%

%%%%%%%%%%%%%%%%%%%%%%%%%%%%%%%%%%%%%%
\subsubsection{Franklin's Lida model }%{Modèles Lida}
%%%%%%%%%%%%%%%%%%%%%%%%%%%%%%%%%%%%%%

The LIDA model - Learning Intelligent Distribution Agent - \cite{franklin_lida_2014} was developed by Franklin up until 2013 by adding the learning factor to the IDA model dating back to the 2000s. 
The cognitive cycle proposed by Franklin relies on a degree of consciousness attributed to information. The cycle begins with a phase of \textit{understanding}, followed by \textit{consciousness}, and concludes with \textit{action selection}, incorporating learning.

Starting with sensory effectors, we can note the creation of the \textit{workspace} module, based on constructing a representation of the current situation (\textit{Current Situational Model}). This information processing step is reminiscent of working memory in more classical models but highlights the design of an internal representation of the situation, enriched by long-term memory.

The workspace develops a dynamic and specific internal representation of the agent based on the current situation. The global workspace is more general, linked to consciousness and information dissemination.

With numerous modules and connections, the model is relatively complex. It integrates several cognitive mechanisms inspired by neuroscience, such as attention, long-term memory, and decision-making. One of the most innovative aspects of LIDA is its ability to simulate consciousness by attributing a degree of awareness to the various pieces of information processed. This mechanism allows the agent to focus on the most relevant stimuli and make decisions more suited to the situation. 

In welding example, auditory cues are subconsciously processed. Significant deviations from the expected welding sound can trigger operator awareness

Learning is described as a continuous process based on adapting internal representations according to experiences, which is essential in dynamic situations.

The cycle end with the realization of an action (motor), influencing the state of the situation. Coupling with the learning and impact of the global workspace, the LIDA model depict the dual regulation of activity.

\begin{figure}
    \vspace{-0,3 cm}
    \centering
    \includegraphics[width=0.99\linewidth]{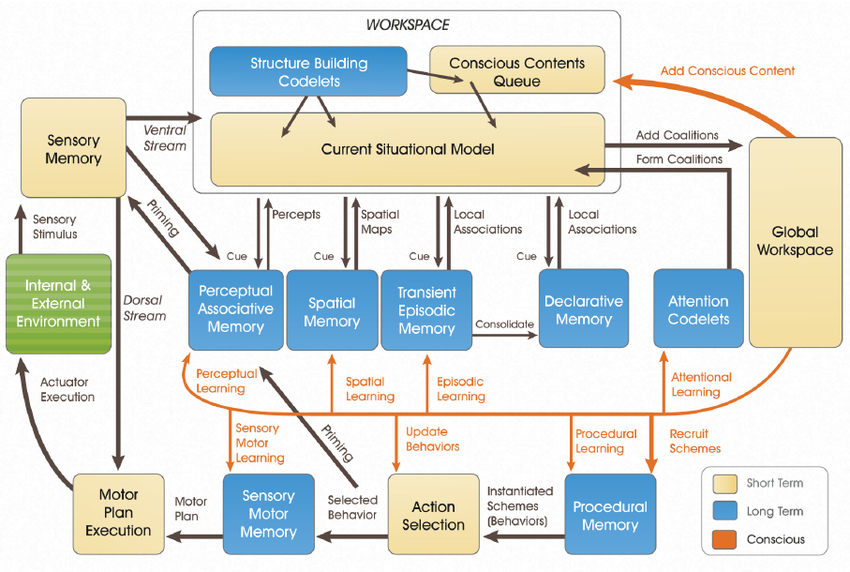}
    \caption{Representative LIDA model diagram \cite{franklin_lida_2014}}%{Schéma représentatif du modèle LIDA \cite{franklin_lida_2014}}
    \label{fig:LIDA}
    \vspace{-0,3 cm}
\end{figure}

%%%%%%%%%%%%%%%%%%%%%%%%%%%%%%%%%%%%%%
\subsubsection{Baddeley's multiple component model}%{Modèle de Baddeley}
%%%%%%%%%%%%%%%%%%%%%%%%%%%%%%%%%%%%%%
This model describes working memory as an autonomous system. It was inspired by the work of Norman and Shallice to develop its central executive system. The multiple component model has undergone several major developments since 1986, with a more recent update in 2021 focusing on the role of attention. The 1986 model is the most cited, but the 2000 version is more refined and does not appear to have had any major updates since then. 

Initially, the model was composed of three modules. The phonological loop treating the verbal and auditory information, the visuo-spatial sketchpad treating  visual and spatial information, and the central executive coordinating both modules and level of attention.

\begin{figure}
    \vspace{-0,3 cm}
    \centering
    \includegraphics[width=0.75\textwidth]{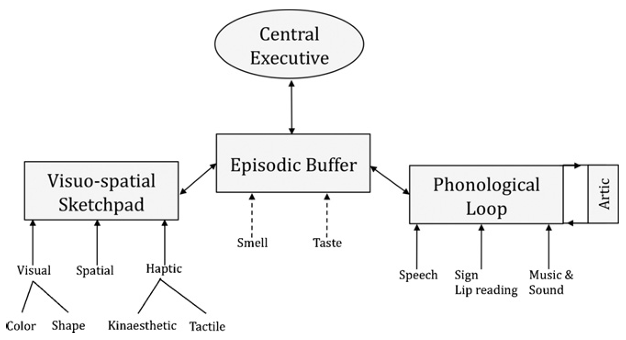}
    \caption{A revised model of working memory by Baddeley \cite{baddeley_binding_2011}}%{Evolution du modèle de Baddeley, extrait de Barouillet \cite{barrouillet_memoire_2022}}
    \label{fig:Baddeley_Recent}
    \vspace{-0,3 cm}
\end{figure}

Norman and Shallice's model is based on schemes (initially \textit{schemeta}) that govern all of our behaviors (both motor and cognitive processes), from the simplest to the most complex. This system activates several schemes in parallel and includes a conflict management module. The activation of schemes, their inhibition, and this conflict management echo Pascual Leone's model.

The cognitive cycle here begins with the perceptual system, which then transmits the information to the database. The information ultimately reaches the scheme control unit, which interacts with the supervisory attentional system and a conflict management module to eventually reach the effector system.

In a first approach, the management of perceived information seems underdeveloped, particularly in terms of attention and information selection. Similarly, the management of knowledge with the database remains unclear. This architecture serves as an interesting basis, demonstrating a cognitive cycle and the management of automatism. Taken up by Baddeley, these schemes are controlled by the central executive, while the perceptual system is separated into a phonological loop and a visuospatial sketchpad. A buffer is added between the executive and the two previous modules, reminiscent of working memory.

As presented by Barrouillet and Camos \cite{barrouillet_memoire_2022} (see figure \ref{fig:Baddeley_Recent}), Baddeley's model has evolved significantly, highlighting the multiple sources of sensory information and describing in more detail the role of the central executive. The latter is based on the choice of the action to be performed, which occurs automatically once triggered, echoing the pattern matching of ACT-R.

The visuo-spatial sketchpad effectively models welding, encompassing both visual weld pool perception and the proprioception required for torch manipulation. This model facilitates the correlation between these aspects, aligning well with observed reality.

%%%%%%%%%%%%%%%%%%%%%%%%%%%%%%%%%%%%%%
\subsubsection{Hollnagel's COCOM cognitive model}%{COCOM de Hollnagel}
%%%%%%%%%%%%%%%%%%%%%%%%%%%%%%%%%%%%%%

COCOM is an acronym for \qquote{Contextual Cognitive Model} developed by Hollnagel \cite{hollnagel_context_1998}. Although not presented as a cognitive architecture, this cognitive model proposes a set of functional processes and information management techniques to analyze and predict behavior. According to Hoc and Amalberti \cite{hoc1999analyse}, the term architecture refers to the framework, meaning the elementary cognitive activities, while the model serves to describe \qquote{by comparison between situations, both the invariants of the activity for the class of considered situations or, on the contrary, the differences}. Hollnagel’s model falls into the second category, and we observe the definition of Vergnaud's scheme in Hoc and Amalberti's description.

Hollnagel aims to develop a model of human behavior that includes the competencies engaged in a human-machine environment. For him, a cognitive model must be able to: \qquote{provide better predictions of the developments that may occur, hence help the design of specific solutions and to assess possible risks}. Thus, a good cognitive control model should not describe a sequence of actions but rather a way to explain how these actions were planned and chosen.

The COCOM model, therefore, focuses on the analysis of behavior and task planning. It is based on three major axes: \qquote{skill, control, and constructs} (p.13). It defines control as the way a subject performs actions and how those choices are made (p.14).

\begin{table}[!ht]
    \centering
    \caption{Main parameters of the COCOM model, extract from \cite[p.22]{hollnagel_context_1998}}%{Paramètres principaux du modèle COCOM, extrait de \cite[p.22]{hollnagel_context_1998}}
    \label{tab:COCOM}
    \begin{tabularx}{\textwidth}{|p{2.4cm}|p{2.3cm}|p{2.3cm}|p{2.3cm}|p{2.3cm}|}
        \hline
     & \textbf{Number of goals} & \textbf{Subjectively available time} & \textbf{Choice of next action} & \textbf{Evaluation of outcome}  \\
     \hline
     \textbf{Strategic} & Several & Abundant & Prediction based & Elaborate \\
     \hline
     \textbf{Tactical \newline (attended)} & Several (limited) & Limited, but adequate & Plan based & Normal \\
     \hline
     \textbf{Tactical \newline (unattended)} & Several (limited) & More than adequate & Plan based, but unreflective & Normal, but imprecise \\
     \hline
     \textbf{Opportunistic} & One, or two (competing) & Short or inadequate & Association based & Cursory \\
     \hline
     \textbf{Scrambled} & One & Very limited & Random & Rudimentary \\
     \hline
    \end{tabularx}
\end{table}

Cognitive control largely corresponds to task planning in the short term. This planning depends on the context (in other words, the situation) and the subject's knowledge (both explicit and implicit). Similarly, the evolution of the situation provides feedback that the subject can perceive to analyze the dependencies between multiple actions. In the COCOM model, cognitive control can occur at five levels (scrambled, opportunistic, tactical (subdivided into attended and unattended), strategic), ranging from actions that appear irrational (Scrambled control mode) to enlightened decision-making with long-term goals (strategic control mode). 

The tactical control mode is heavily inspired by Rasmussen's SRK (Skills Rules Knowledge) model \cite{rasmussen_skills_1983}, as it follows known procedures or rules (R level of Rasmussen). These control modes are selected based on the available knowledge, the stress of the situation as allowed time response and consequences of actions, desired resource economy, and the objective's complexity, which varies depending on the task's nature and the subject's understanding of it. 

Thus, the choice of control mode is summarized by two parameters: perceived time (subjective) and familiarity with the situation. A familiar situation will rely more on automatism, leading to behavior closer to a tactical mode. Conversely, an exceptional situation will necessitate a strategic mode.

The third major axis mentioned by Hollnagel, constructs, refers to concepts, that is, the set of an individual's beliefs about the current situation. \qquote{Some constructs are like hypotheses, some are more like beliefs, and some represent general or “universal” knowledge. Constructs, as well as competence, can change as a consequence of learning} (p.17). Skills, unlike \emph{constructs}, always encompass the ability to achieve something, but these two terms are intrinsically linked. This definition reminds the operational invariants of scheme.

The four categories of control are not absolute; they provide a general idea of how the operator interprets a situation, understands its state, anticipates the consequences of previous actions, and evaluates the resources available, thereby shaping their competencies. Similarly, the model should help explain the conceptualization required to communicate actions. If we understand all these parameters, we can then predict why the situation is under control and how, if applicable, a loss of control can be observed. \qquote{Constructs are similar to the schemeta of Neisser (1976) in that they serve as the basis for selecting actions and interpreting information. (p.17)}.

The COCOM model, specifically, includes the following elements:  
\begin{itemize}  
    \item Number of objectives, closely related to mental load (according to Hollnagel),  
    \item Perception of time,  
    \item Action selection: it can be based on a strategic prediction with thoughtful and expected consequences or be random. This choice is linked to the operator's competencies and \emph{constructs}, which enable or hinder their ability to understand the characteristics of the action to achieve their objective over a short or long term.  
    \item Result evaluation, which can be either elaborate or rudimentary.  
\end{itemize}  

\begin{figure}
    \vspace{-0,3 cm}
    \centering
    \fbox{\includegraphics[width=0.65\textwidth]{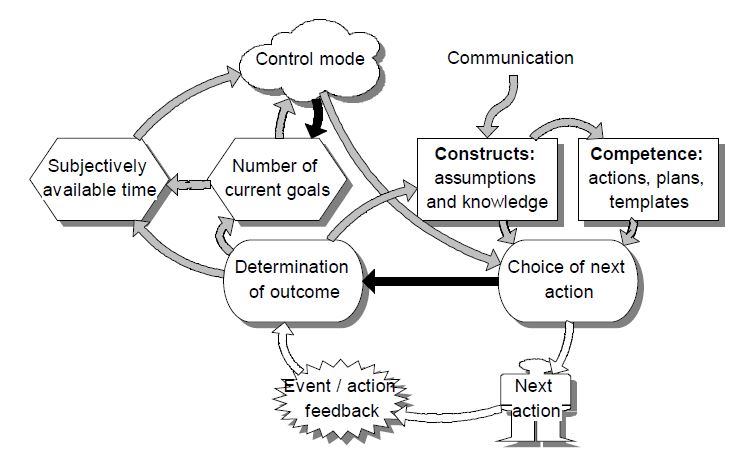} }
    \caption{Functional relations of the COCOM model parameters, extract from \cite[p.22]{hollnagel_context_1998}}%{Relations fonctionnelles des paramètres du modèle COCOM, extrait de \cite[p.22]{hollnagel_context_1998}}
    \label{fig:RelationsCOCOM}
\end{figure}

These various elements provide a cognitive control model summarized in Fig. \ref{fig:RelationsCOCOM} and Table \ref{tab:COCOM}.

This model notably enables the linkage of performance to context and the control exercised by the operator. However, it is important to note that such models are mere simplifications of all cognitive processes since there is neither a single type of control nor a simple context but rather a myriad of parameters describing these two components. The advantage of COCOM lies in its flexibility in viewing cognitive control through multiple variables with various states.  

The plurality of dimensions abstracting cognitive control led Amalberti and Hoc \cite{hoc_cognitive_2007} to develop their cognitive control model based on the level of abstraction of the data used for control and the origin of these data. The objective was to integrate the two types of models from Rasmussen and Hollnagel.  

Like Hollnagel, Hoc and Amalberti \cite{hoc_cognitive_2007} emphasize the control of the situation by developing the principle of a sense of mastery over the situation associated with the notion of acceptable performance.  

In his 2001 article, Amalberti \cite{amalberti_maitrise_2001} defines cognitive control as :  \qquote{any supervisory activity, internal to cognition, aimed at ensuring and verifying the proper use of cognitive abilities in terms of both intensity and temporal sequencing, in order to achieve the goal or goals set by the individual}.  

This definition of cognitive control by Hoc and Amalberti is thus close to that of Hollnagel; they consider cognitive control as making a choice in resource allocation to achieve a goal. That is, the subject, consciously or unconsciously, will organize their cognitive abilities and apply them with a certain intensity to achieve a sense of mastery over the situation.  

A skilled welder, observing a deforming weld pool, can anticipate its evolution. Due to time constraints and process familiarity, their control could often resides in the tactical domain. However, when presented with a novel material, the welder transitions to strategic control, demanding high attention for material selection and process adaptation

%%%%%%%%%%%%%%%%%%%%%%%%%%%%%%%%%%%%%%%%%%%%%%%%%%%
\subsection{Point of interest in cognitive architecture}
%%%%%%%%%%%%%%%%%%%%%%%%%%%%%%%%%%%%%%%%%%%%%%%%%%%
As mentioned earlier, we aim to model skills as a foundation for developing an effective robotization strategy. To do that, we want to have a representation as block diagrams, incorporating concepts from cognitive architectures. Giving an overview of the cognitive structure can facilitate the dialogue with roboticist about input and output needed for there machine. A cognitive architecture model can provide a module-based structure to serialize the cognitive cycle and address the limitations of the conceptual field theory. 

Although we do not aim to \qquote{choose} a specific architecture but rather to draw inspiration from the overall concept, it is relevant for us to explore one of these architectures in detail. To address the limitations of the scheme, we are looking for simple architectures. While the LIDA architecture is highly comprehensive, its complexity makes it less practical to use. The ACT-R model, in particular, provides a framework that is easy to implement and maintain. The COCOM model offers a set of functional relationships between the model's parameters (see Figure \ref{fig:RelationsCOCOM}); these relationships are not overly numerous and allow for quick understanding of the model.

Regulation of activity are crucial in dynamic situations. In our representation, we aim to emphasize the aspect of dual regulation of activity. Some architectures, such as LIDA or the return of pattern matching in ACT-R, highlight this aspect, whereas it is less present in COCOM or Laird's models.

Another criterion, considered so far, is the established reputation of the proposed models within the scientific community. Dating back to the 1980s and having undergone significant evolution, SOAR and ACT-R have been successfully applied in numerous contexts, an important criterion for task modeling.

There exists a wide variety of cognitive architectures, often specific to a domain or task resolution. Attempts to develop standard models, such as those by Laird or Franklin, aim to propose a model that is sufficiently precise to describe cognitive functioning while being broad enough to adapt to all situations. We have chosen to focus on a limited number of models, as we do not intend to develop a new one but rather to draw from existing representations to skill modelisation. Skills, particularly in complex problem-solving tasks or dynamic situations, relies on perception-action loops that are well-described in cognitive architectures. Therefore, focusing on cognitive architectures that have been successfully applied in various contexts (such as SOAR and ACT-R models) seemed sufficient to us.

The temporal delay and the non-systematic compliance feedback in welding introduce a unique learning dimension. Knowledge development is highly personal, leading to strong inter-individual variations. The use of knowledge, particularly implicit knowledge, and learning factors such as emotions, motivation, and attention levels are crucial to understanding skill. However, they are not essential for modeling it. The LIDA architecture includes all these notions, while COCOM considers some of these elements, such as determining the mode of control based on stress or perceived time, for example.

Numerous comparisons are possible between the SOAR and ACT-R models. Additionally, the COCOM or LIDA models can align with the described cognitive cycle and share comparable elements. We identify the starting point with sensory effectors on the one hand and the comparison between knowledge and the situation. We can see a similarity with scheme's inferences: if such parameters, then such rules of action (see definition given earlier). The COCOM model, with its five levels of decision-making, is particularly interesting for conceptualizing skill and its implementation. It could be linked to Anderson's pattern matching or the access between long-term memory and working memory in SOAR/Standard.We propose, in Table \ref{tab:comparison}, a comparison of the various models.

\begin{table}[!ht]
    \centering
    \caption{Comparison of different models}
    \begin{tabular}{|c|p{4.8cm}|p{4.8cm}|}
        \hline
        Model & Advantages & Disadvantages \\
        \hline
        Standard & Simplicity & Difficulty in integrating the dual regulation of activity and explaining intra-individual variations. \\
        \hline
        LIDA & Considers emotions and attention, explains intra-individual variations, more precise learning. & Complex model \\
        \hline
        SOAR & More precise than the standard model. Numerous succesful applications & scheme less easily integrated compared to the ACT-R model, less flexible than the standard model. \\
        \hline
        ACT-R & Simple structure, numerous successful applications, easy scheme integration, visually represents dual regulation of activity, capable of explaining intra-individual variations. &  \\
        \hline
    \end{tabular}
    \label{tab:comparison}
\end{table}

Building upon the strengths of ACT-R and additions of COCOM and LIDA, which aligns with our criteria for simplicity, successful applications, scheme integration, dual activity regulation, and explanation of intra-individual variations, we aim to develop a skill model inspired by cognitive architectures. Some refinement are needed to go through ACT-R's limitations such as the assumption of complete rationality in decision-making and the pattern-matching concept's difficulty in explaining expert errors. Therefore, we propose a block diagram representation of our skill model, drawing inspiration from ACT-R and incorporating insights from other relevant cognitive theories to address these limitations.

%%%%%%%%%%%%%%%%%%%%%%%%%%%%%%%%%%%%%%%%%%%%%%%%%%%
\section{Presentation of our model}
\begin{figure}
    \vspace{-0,3 cm}
    \centering
    \includegraphics[width=\textwidth]{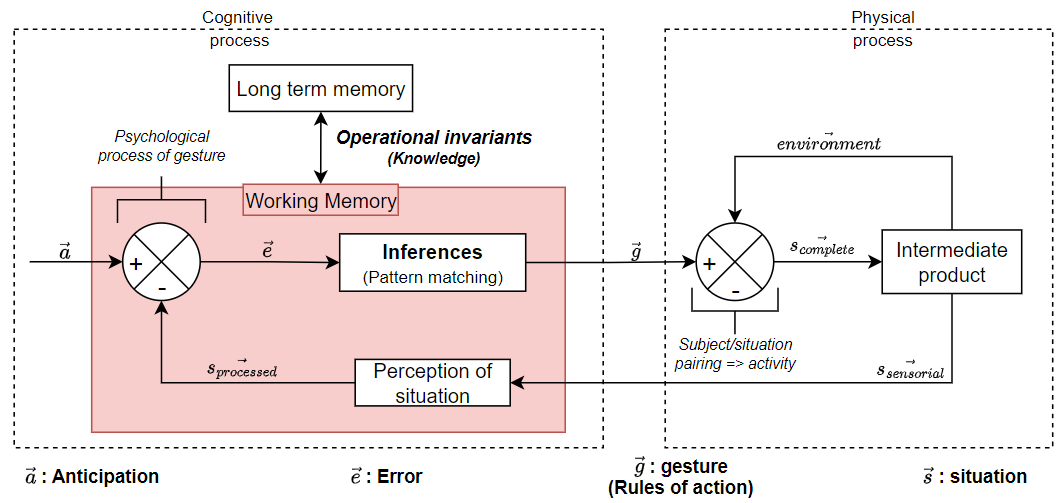}
    \caption{Skill model integrating Vergnaud's scheme and cognitive architectures}
    \label{fig:Modele_Competence}
    \vspace{-0,3 cm}
\end{figure}

%%%%%%%%%%%%%%%%%%%%%%%%%%%%%%%%%%%%%%%%%%%%%%%%%%%
We aim to develop a representation of the skill inspired by the scheme as described by Vergnaud's conceptual field. The model must include the dual regulation of activity and intra-individual variations. Ideally, we seek to understand how knowledge develops over the long term. To define the limits of cognitive functioning, we have drawn inspiration from the ACT-R model. Figure \ref{fig:Modele_Competence} presents our schematic representation of the skill in the context of dynamic situations.

The cognitive cycle begins with the individual’s expectations (vector \(\overrightarrow{a}\) for Anticipation), which are compared to the perception of the situation. The error between the expected ideal situation and the perceived outcome leads to a modification of the action based on the individual's knowledge. The pattern matching is feed with the error between the expectations and the situation perceived. This process is justified by the agent’s operational invariants, which represent the set of beliefs involved in the situation. The gap then prompts an identification of the situation, following an \qquote{if conditions, then action} logic (\textit{i.e.}, inferences). The selected action is then performed, which may involve a motor action (gesture adaptation) or a cognitive action (adjusted expectations, complementary information gathering, etc.), executed either consciously or unconsciously (symbolic or subsymbolic) and with varying degrees of automation. These actions are analogous to schemata described in Baddeley’s model. Cognitive processing influences how we act on the world, either by modifying a gesture or remaining constant, thereby maintaining interaction with the environment, particularly in dynamic situations.  

The execution of the gesture (vector \(\overrightarrow{g}\) for gesture) in the environment produces an intermediate product (\textit{e.g}., in welding, a molten pool that solidifies into a weld bead), inside the complete situation depicted with the vector \(\overrightarrow{s_{complete}}\), which modifies some characteristics of the situation (\textit{e.g.} temperature, shape of the molten pool), represented by a feedback arrow. These modifications also serve as experience for the operator, enabling them to associate actions with changes in the intermediate product. The bidirectional arrow between knowledge and psychological processes indicates the creation of knowledge from the situation and the use of knowledge in action selection. In other words, it represents the adaptation of the scheme to the current situation by the accomodation and assimilation processes. The two arrows originating from the intermediate product represent the dual regulation of activity.  

The intermediate product and the overall situation provide a set of information. The human organism perceives a sample of this information through sensory organs (vector situation \(\overrightarrow{s_{sensorial}}\)). The agent then selects relevant information through a consciousness mechanism inspired by the LIDA model, taking place in the working memory (vector situation \(\overrightarrow{s_{processed}}\)). Additionally, this interpretation is influenced by response time constraints, familiarity with the situation, perceived difficulty, and anticipated consequences. This interpretation ultimately relies on the control levels defined in the COCOM model.  

Vergnaud’s rules of action are present throughout the cycle, as they break down activities into elementary tasks. For example, these include gathering information from each element of the situation, creating a representation of the goals to be achieved, justifying a choice based on a belief, performing the action (itself broken down further), and so on.  

In the context of welding, the expectation vector (\(\overrightarrow{a}\)) includes the shape of the molten pool, the sound, and the torch’s position in the pool. The operative invariants encompass the set of exploited beliefs, such as the rules for adjusting welding intensity, the interactions between variables, and the prediction of changes in the pool caused by specific actions (\textit{e.g}., reducing the pool size and shifting the torch’s position in the pool if the welding speed is increased). The intermediate product is the current state of the molten pool. The environment consists of numerous variables, among which temperature and gravity (depending on the welding position) are particularly significant. The human body detects a set of elements through sensory organs influenced by the environment (\textit{e.g}., the eyes are behind a welding visor). The cognitive system then selects the important information to create a result that is compared to the expectations. 

This cycle continues until the welder gesture is to interrupts the electric arc. This interruption halts the melting process, allowing the weld pool to cool and solidify, thereby concluding the welding operation

It is, however, interesting to consider whether the operator forms a mental representation stored in memory or continuously updates it. While the proposed model accommodates both cases, it necessitates further investigation into task parallelization and working memory management.

%%%%%%%%%%%%%%%%%%%%%%%%%%%%%%%%%%%%%%%%%%%%%%%%%%%
\section{Conclusions and future work}
%%%%%%%%%%%%%%%%%%%%%%%%%%%%%%%%%%%%%%%%%%%%%%%%%%%

This work proposes a novel cognitive model of skill, integrating concepts from cognitive psychology, including scheme theory, cognitive architectures, dual regulation of activity, contextual cognitive models, and the perception-action loop.  By building upon these established frameworks, our approach offers a visualization of the dual regulation of activity within perception-action loops, providing a precise representation of the cognitive processes underlying human skill execution in complex tasks. This model not only opens new perspectives for developing human-inspired robotization strategies, enabling robots to better collaborate with and augment human capabilities, but also provides a valuable tool for skill evaluation and training, particularly within professional training contexts. For instance, the model could be used to identify specific areas where trainees struggle and develop training programs to address those weaknesses.

This model, while promising, requires further development such as refining the temporal delay parameters, managing the realization of automatic action (schemata), and investigating the impact of variables like workload and stress on performance (like shown in COCOM model). Model validation against empirical data from expert performance are needed to predict capabilities, and ultimately contribute to a robust, global skill model applicable across domains, advancing human-machine collaboration in Industry 5.0.

\bibliographystyle{splncs04}
\bibliography{bib-list}
\end{sloppypar}
\end{document}